\documentclass[11pt, a4paper]{article}
\usepackage{amsmath,amssymb,amsthm,amsfonts,bbm}
\usepackage{color,setspace,fullpage,algorithm,algorithmic}
\usepackage{ccfonts,eulervm} 
\usepackage[T1]{fontenc}

\theoremstyle{plain}
\newtheorem{thm}{Theorem}

\theoremstyle{definition}

\newcommand{\ip}[2]{\left\langle #1,#2 \right\rangle}
\newcommand{\bydef}{\stackrel{\bigtriangleup}{=}}
\newcommand{\expect}[1]{\mathbb{E}\left[{#1}\right]}

\newcommand{\prob}[1]{\mathbb{P}\left[{#1}\right]}

\newcommand{\given}{\; \big\vert \;}
\newcommand{\norm}[1]{\left\| #1 \right\|}

\begin{document}
\title{Thompson Sampling for Online Learning\\with Linear Experts}
\author{Aditya Gopalan\\Technion, Haifa, Israel\\{\tt
    aditya@ee.technion.ac.il}}
\maketitle

\begin{abstract}
  In this note, we present a version of the Thompson sampling
  algorithm for the problem of online linear generalization with full
  information (i.e., the experts setting), studied by Kalai and
  Vempala, 2005. The algorithm uses a Gaussian prior and time-varying
  Gaussian likelihoods, and we show that it essentially reduces to
  Kalai and Vempala's Follow-the-Perturbed-Leader strategy, with
  exponentially distributed noise replaced by Gaussian noise. This
  implies sqrt(T) regret bounds for Thompson sampling (with
  time-varying likelihood) for online learning with full information.
\end{abstract}

\section{Setup}
Consider the full-information linear generalization setting, similar
to the one studied by Kalai and Vempala \cite{KalaiV05}. We can
select, at each time $t \geq 1$, a decision $d_t$ from an action set
$\mathcal{D} \subset \mathbb{R}^n$. Following the $t$-th decision
$d_t$, we get to observe $s_t \in \mathcal{S} \subset \mathbb{R}^n$
and receive a reward of $\ip{d_t}{s_t}$. The goal is to maximize the
total reward $\sum_t
\ip{d_t}{s_t}$. \\

As shorthand we will write $S_t$ for the vector $s_1 + s_2 + \ldots +
s_t$, and $x_i$ for the $i$-th coordinate of a vector $x$. Throughout,
$I_n$ and $\mathbbm{1}_n$ denote the identity matrix and all-ones
vector in dimension $n$ respectively. \\

Consider the Thompson Sampling algorithm TSG($\epsilon$), with
$\epsilon > 0$, and Gaussian prior and likelihood (Algorithm
\ref{alg:tsgauss}).

\begin{algorithm}[htbp]
  \caption{TSG($\epsilon$)}
  \label{alg:tsgauss}
  {\bf for} $t = 1, 2, 3, \ldots$
  \begin{enumerate}
  \item Assume that $\{s_{k}\}_{k < t}$ are independent and
    identically distributed $\mathcal{N}(\mu,\frac{1}{\sqrt{\epsilon
        (t-1)}}I_n)$ samples, where $\mu$ follows the prior distribution
    $\mu \sim \mathcal{N}(0,\frac{1}{\epsilon}I_n)$. Draw $\theta_t
    \in \mathbb{R}^n$ from the posterior distribution $\prob{\mu
      \given \{s_k\}_{k < t}}$.
  \item Play $d_t = \arg \max_{d \in \mathcal{D}} \ip{d}{\theta_t}$.
  \end{enumerate}
  {\bf end for}
\end{algorithm}

By standard results, upon observing iid standard normal samples $x_1,
\ldots, x_{t-1}$ distributed as $\mathcal{N}(\mu,\sigma^2)$, with
nonrandom variance $\sigma^2$ and prior $\mu \sim
\mathcal{N}(\mu_0,\sigma_0^2)$, the posterior distribution of the mean
$\mu$ is again Gaussian with mean $\frac{\sigma_0^2\overline{x} +
  \frac{\sigma^2}{t-1}\mu_0 }{\sigma_0^2 + \frac{\sigma^2}{t-1}}$ and
variance $\left(\frac{1}{\sigma_0^2} +
  \frac{t-1}{\sigma^2}\right)^{-1}$. In our case, at time $t \geq 2$,
\begin{align*}
  \prob{\theta_t \given \{s_k\}_{k < t}} &\sim \mathcal{N}\left(
    \frac{\frac{1}{\epsilon}}{\frac{1}{\epsilon} + \frac{1}{\epsilon
        (t-1)^2}} \times \frac{S_{t-1}}{t-1}, \frac{1}{\epsilon + \epsilon(t-1)^2}\right) \\
  \Rightarrow \; \prob{\left(t-1 + \frac{1}{t-1} \right)  \theta_t \given \{s_k\}_{k < t}} &\sim
  \mathcal{N}\left(S_{t-1} , \epsilon^{-1} \left(1+\frac{1}{(t-1)^2}\right) \right) \\
  &\sim S_{t-1} + \mathcal{N}\left(0,\epsilon^{-1} \left(1+\frac{1}{(t-1)^2}\right)\right).
\end{align*} 
Thus, the TSG algorithm perturbs the aggregate `state' $S_{t-1}$ seen so far
with Gaussian noise, and takes the best decision for this perturbed
state. This is akin to the Follow-the-Perturbed-Leader (FPL) strategy
developed by Kalai and Vempala \cite{KalaiV05}, and we apply their
techniques to provide regret bounds for TSG that hold over all
sequences $s_1, s_2, \ldots $ in $\mathcal{S}$. Our result involves
the following parameters:
\begin{align*}
  D &\bydef \sup_{d,d' \in \mathcal{D}} \norm{d-d'}_1, \quad\quad R \bydef \sup_{d \in
    \mathcal{D},s \in \mathcal{S}} |\ip{d}{s}|, \quad\quad A_1 \bydef \sup_{s \in
    \mathcal{S}} \norm{s}_1, \quad\quad A_2 \bydef \sup_{s \in
    \mathcal{S}} \norm{s}_2.
\end{align*}
As usual, for a sequence of states $s_1, s_2, \ldots, s_T$, we define
the \emph{regret} $R^A(T)$ of a strategy $A$ to be the difference
between the reward earned by $A$ on the sequence and the reward earned
by the best fixed decision in hindsight:
\[ R^A(T) \bydef \\sup_{d \in \mathcal{D}} \sum_{t=1}^T \ip{d}{s_t}  \sum_{t=1}^T \ip{d_t^A}{s_t}.  \]

\begin{thm}
  The expected regret of TSG($\epsilon$) satisfies
  \[ \expect{R^{TSG(\epsilon)}(T)} \leq \sqrt{\epsilon} R A_2 K_{2,n}
  T + \frac{\epsilon R A_2^2 T}{2} +
  \frac{2DK_{\infty,n}}{\sqrt{\epsilon}}, \] where $K_{2,n}$ and
  $K_{\infty,n}$ are positive constants that depend only on $n$.
\end{thm}

{\bf Note:} Setting $\epsilon = \frac{1}{T}$ implies an expected
regret of $O(\sqrt{T})$.

\begin{proof}
Let us introduce the notation $M(x) \bydef \arg\max_{d \in D}
\ip{d}{x}$. TSG chooses the decision $M(S_{t-1}+p_t)$ at time $t$,
where $p_t \sim \mathcal{N}\left(0,\epsilon^{-1} \left(1+q_t\right)
\right)$, $q_t = \frac{1}{(t-1)^2}$. 

First, an application of Lemma 3.1 in \cite{KalaiV05} gives that for
any state sequence $s_1, s_2, \ldots$, $T > 0$ and vectors $p_0 = 0,
p_1, \ldots, p_T$,
\begin{align}
  \ip{M(S_T)}{S_T} &\leq \sum_{t=1}^T \ip{M(S_t + p_t)}{s_t} +
  D\sum_{t=1}^T \norm{p_t - p_{t-1}}_\infty. \label{eqn:basicbd}
\end{align}
Next, observe that the expected reward is unchanged if for each $t >
1$, $p_t = p_1\sqrt{1+q_t}$. For such a noise sequence,
\begin{align}
  \norm{p_t - p_{t-1}}_\infty &= \norm{p_1}_\infty\cdot \left|\sqrt{1+q_t} - \sqrt{1+q_{t-1}}\right| \nonumber \\
  &\leq \norm{p_1}_\infty\cdot \left|\left(\sqrt{1+q_t}\right)^2 - \left(\sqrt{1+q_{t-1}}\right)^2 \right| \nonumber \\
  &= \norm{p_1}_\infty\cdot \left|q_t - q_{t-1} \right| \nonumber \\
  \Rightarrow \; \sum_{t=2}^T \norm{p_t - p_{t-1}}_\infty &\leq \norm{p_1}_\infty\sum_{t=2}^T  \left|q_t - q_{t-1} \right| \nonumber \\
  &= \norm{p_1}_\infty\sum_{t=2}^T  \left(\frac{1}{(t-1)^2} - \frac{1}{t^2} \right) \nonumber \\
  &\leq \norm{p_1}_\infty. \label{eqn:noisebd}
\end{align}
TSG earns reward $\ip{M(S_{t-1} + p_t)}{s_t}$ at each time $t$, and
the best possible reward in hindsight over the entire time horizon $1,
2, \ldots, T$ is $\ip{M(S_T)}{S_T}$, so in order to bound the regret
of TSG using (\ref{eqn:basicbd}), it remains to bound the expectation of
the difference $\ip{M(S_t + p_t)}{s_t} - \ip{M(S_{t-1} +
  p_t)}{s_t}$. Let $\epsilon_t^{-1} \bydef \epsilon^{-1}(1+q_t)$, and let
$d\nu_a(\cdot)$ be Gaussian measure on $\mathbb{R}^n$ with mean $0$
and variance $a^{-1}I_n$. Observe that
\begin{align*}
  \expect{\ip{M(S_{t-1} + p_t)}{s_t}} &= \int_{x \in \mathbb{R}^n} \ip{M(S_{t-1} + x)}{s_t} \; d\nu_{\epsilon_t}(x) \\
  &= \int_{y \in \mathbb{R}^n} \ip{M(S_{t-1} + s_t + y)}{s_t} \; d\nu_{\epsilon_t}(y + s_t) \\
  &= \int_{y \in \mathbb{R}^n} \ip{M(S_{t} + y)}{s_t} \; d\nu_{\epsilon_t}(y + s_t) \\
  &= \int_{y \in \mathbb{R}^n} \ip{M(S_{t} + y)}{s_t} e^{\frac{\epsilon_t}{2}\left(\norm{y}_2^2 - \norm{y+s_t}_2^2\right)} \; d\nu_{\epsilon_t}(y).
\end{align*}
Thus, we can write
\begin{align*}
  & \expect{ \ip{M(S_{t} + p_t)}{s_t} - \ip{M(S_{t-1} + p_t)}{s_t}  } \\
  &= \int_{y \in \mathbb{R}^n} \ip{M(S_{t} + y)}{s_t} \left[1 -
      e^{\frac{\epsilon_t}{2}\left(\norm{y}_2^2 - \norm{y+s_t}_2^2\right)}
    \right]
    \; d\nu_{\epsilon_t}(y) \\
  &=  \int_{z \in \mathbb{R}^n} \ip{M(S_{t} + \epsilon_t^{-\frac{1}{2}}z )}{s_t} \left[1 -
    e^{\frac{1}{2}\left(\norm{z}_2^2 - \norm{z+s_t\sqrt{\epsilon_t}}_2^2\right)}
  \right] \; d\nu_{1}(z) \\
  &= \int_{z \in \mathbb{R}^n} \ip{M(S_{t} + \epsilon_t^{-\frac{1}{2}}z )}{s_t} \left[1 -
    e^{\frac{1}{2}\left( -2\ip{z}{s_t\sqrt{\epsilon_t}} - \epsilon_t \norm{s_t}_2^2  \right)}
  \right] \; d\nu_{1}(z) \\
  &\leq \int_{z \in \mathbb{R}^n} \ip{M(S_{t} + \epsilon_t^{-\frac{1}{2}}z )}{s_t} \left[1 -
    e^{-\sqrt{\epsilon_t}\norm{s_t}_2 \norm{z}_2 -\frac{\epsilon_t A_2^2}{2} }
  \right] \; d\nu_{1}(z) \\
  &\quad \mbox{(Cauchy-Schwarz, and assuming that $\ip{d}{s} \geq 0$ $\forall d \in \mathcal{D}, s \in \mathcal{S}$)} \\
  &\leq \int_{z \in \mathbb{R}^n} \ip{M(S_{t} + \epsilon_t^{-\frac{1}{2}}z )}{s_t} \left[\sqrt{\epsilon_t}\norm{s_t}_2 \norm{z}_2 + \frac{\epsilon_t A_2^2}{2}\right] \; d\nu_{1}(z) \\
  &\quad \mbox{(since $1-e^{-x} \leq x$)}  \\
  &\leq \sqrt{\epsilon_t} R A_2 K_{2,n} + \frac{\epsilon_t R A_2^2}{2} \leq \sqrt{\epsilon} R A_2 K_{2,n} + \frac{\epsilon R A_2^2}{2},
\end{align*}
where $K_{p,n} \bydef \int_{z \in \mathbb{R}^n} \norm{z}_p \; d\nu_{1}(z)$
for $p \geq 1$. Combining the above with (\ref{eqn:basicbd}) and
(\ref{eqn:noisebd}) and summing over $1, 2, \ldots, T$ gives
\begin{align*}
  \expect{\ip{M(S_T)}{S_T}} - \expect{\sum_{t=1}^T
    \ip{d_t^{TSG}}{s_t}} &\leq \sqrt{\epsilon} R A_2 K_{2,n} T +
  \frac{\epsilon R A_2^2 T}{2} + \frac{2DK_{\infty,n}}{\sqrt{\epsilon}},
\end{align*}
completing the proof. 
\end{proof}

\bibliographystyle{ieeetr}

\end{document}